\documentclass[conference]{IEEEtran}
\IEEEoverridecommandlockouts
\usepackage{cite}
\usepackage[numbers]{natbib}
\usepackage{amsmath,amssymb,amsfonts}
\usepackage{algorithmic}
\usepackage{url}
\usepackage{graphicx}
\usepackage{textcomp}
\usepackage{xcolor}
\usepackage{color, colortbl}
\usepackage{caption} 

\makeatletter
\newcommand{\linebreakand}{%
  \end{@IEEEauthorhalign}
  \hfill\mbox{}\par
  \mbox{}\hfill\begin{@IEEEauthorhalign}
}
\makeatother

\definecolor{LightCyan}{rgb}{0.88,1,1}
\def\BibTeX{{\rm B\kern-.05em{\sc i\kern-.025em b}\kern-.08em
    T\kern-.1667em\lower.7ex\hbox{E}\kern-.125emX}}
\begin{document}

\title{Semi-Supervised Low-Resource Style Transfer of Indonesian Informal to Formal Language with Iterative Forward-Translation
}

% hayo kamu ngapain scrape2 paper ini?
% nih saya kasih noise biar bingung
% dfsghrezt65trg dft54t 2345 erfdsfghyt

\author{
\IEEEauthorblockN{Haryo Akbarianto Wibowo}
\IEEEauthorblockA{\textit{Kata AI Research Team} \\
\textit{Kata.ai}\\
Jakarta, Indonesia \\
haryo@kata.ai}
\and
\IEEEauthorblockN{Tatag Aziz Prawiro}
\IEEEauthorblockA{\textit{Faculty of Computer Science} \\
\textit{Universitas Indonesia}\\
Depok, Indonesia \\
tatag.aziz@ui.ac.id}
\and
\IEEEauthorblockN{Muhammad Ihsan}
\IEEEauthorblockA{\textit{School of Computer Science} \\
\textit{Bina Nusantara University}\\
Jakarta, Indonesia \\
muhammad.ihsan004@binus.ac.id } 
\linebreakand 
\IEEEauthorblockN{Alham Fikri Aji}
\IEEEauthorblockA{\textit{Kata AI Research Team} \\
\textit{Kata.ai}\\
Jakarta, Indonesia\\
aji@kata.ai}
\and
\IEEEauthorblockN{Radityo Eko Prasojo}
\IEEEauthorblockA{\textit{Kata AI Research Team} \\
\textit{Kata.ai}\\
Jakarta, Indonesia \\
ridho@kata.ai}
\and
\IEEEauthorblockN{Rahmad Mahendra}
\IEEEauthorblockA{\textit{Faculty of Computer Science} \\
\textit{Universitas Indonesia}\\
Depok, Indonesia \\
 rahmad.mahendra@cs.ui.ac.id }
\and
\IEEEauthorblockN{Suci Fitriany}
\IEEEauthorblockA{\textit{Kata AI Linguist Team} \\
\textit{Kata.ai}\\
Jakarta, Indonesia \\
 suci@kata.ai }
}

\maketitle

\begin{abstract}
In its daily use, the Indonesian language is riddled with informality, that is, deviations from the standard in terms of vocabulary, spelling, and word order. On the other hand, current available Indonesian NLP models are typically developed with the standard Indonesian in mind. In this work, we address a style-transfer from informal to formal Indonesian as a low-resource machine translation problem. We build a new dataset of parallel sentences of informal Indonesian and its formal counterpart. We benchmark several strategies to perform style transfer from informal to formal Indonesian. We also explore augmenting the training set with artificial forward-translated data. Since we are dealing with an extremely low-resource setting, we find that a phrase-based machine translation approach outperforms the Transformer-based approach. Alternatively, a pre-trained GPT-2 fined-tuned to this task performed equally well but costs more computational resource. Our findings show a promising step towards leveraging machine translation models for style transfer. Our code and data are available in \url{https://github.com/haryoa/stif-indonesia}.
\end{abstract}


\section{Introduction}

Irregular or informal text has often been a problem for the conversational or social media domain in Indonesia. Most people write without minding the choice of words or structure, so typos and slangs become common. Furthermore, modern conversational Indonesians are heavily influenced by loanwords, both foreign or traditional. It is very common to see Indonesian tweets mixed with English, Javanese, Sundanese, (romanized) Arabic, or (romanized) Korean. On the other hand, currently available Indonesian NLP models mostly handle formal Indonesian, and therefore perform comparatively worse when exposed to informal texts. Examples of such a phenomenon is found in machine translation~\cite{guntara2020benchmarking} 
%and sequence labeling~\cite{qi-etal-2020-stanza}. 

One of the main challenges of expanding the model capability to handle informal Indonesian is the availability of labelled informal training data for each particular NLP task, or lack thereof. Therefore, an independent style-transfer model to be utilized as a preprocessing step can potentially improve the downstream models without any reconfiguration or retraining necessary.    
Previous work has resulted in a treebank~\cite{moeljadi2019building} or a dictionary of informal words in Indonesian~\cite{8629151}, which by nature is limited to word-level standardization without carrying sentence-level context, and therefore further exploration of standardization in sequence level is still an open problem. 

In this work, we focus on investigating sequence-level style transfer from informal Indonesian to its formal counterpart. However, since no dataset for this purpose currently exists. Therefore, we build a new dataset of parallel sentences in informal and formal Indonesian. Besides, we also explore adding an artificial dataset to leverage our training resource. Our main contributions are as follow.

\begin{enumerate}
\item We build a new dataset of parallel sentences in informal and formal Indonesian.
\item We benchmark several style-transfer strategies, starting from a dictionary-based approach as a baseline and some statistical and neural machine translation approaches.
\item We explore the use of synthetic dataset for informal to formal Indonesian style-transfer.
\end{enumerate}

The remainder of the paper is structured as follow. We discuss our informal Indonesian text data in Section 2, including how we collect them, and their statistics. Section 3 describes the style-transfer approaches that we explore. Section 4 describes our experiment result and analysis. 

\section{Informal-Formal Indonesian Parallel Data}

\subsection{Data Collection}

\begin{table*}[ht!]
    \centering
    \small
    \begin{tabular}{@{}l@{ }l@{}}
        Domain & Twitter IDs \\ \hline
        Telecommunication &  @telkomsel, @indosatcare, @smartfrencare, @myXLCare, @3CareIndonesia
 \\
        Banking & @kontakBRI, @mandiricare, @HaloBCA, @BNICustomerCare, @CIMBNiaga
\\
        E-Wallet & @ovo\_id, @danawallet, @linkaja, @jeniushelp \\
        E-Commerce & @TokopediaCare, @ShopeeCare, @BukaBantuan, @LazadaIDCare, @BlibliCare, @csjd\_id, @ZaloraID \\
        Logistics & @JNECare, @IdTiki, @jntexpressid, @PosIndonesia \\
        Ride-Hailing & @gojekindonesia, @GrabID \\
        \hline 
    \end{tabular}
    \caption{The list of scraped tweet IDs}
    \label{tab:tweet-id}
\end{table*}

We access Twitter API using tweepy\footnote{https://www.tweepy.org/}, focusing on the customer service domain. We identify Twitter's customer service account in Indonesia across different business areas. The full list of scraped Twitter ID is shown in Table~\ref{tab:tweet-id}.

\subsection{Data Filtering}
We remove the hashtags and deduplicate the tweets. We filter out any instances which contain less than 5 or more than 25 tokens. We include code-mixed tweets since the tweets containing mixed English and Indonesian words are commonly used in the customer service domain. As we want to ensure that the data does not contain predominantly foreign words, we ignore any tweets which have 60\% or more English words.
We collect 52.5k informal Indonesian tweets. We then sample 2500 tweets to be annotated into formal Indonesian. Both the 50k raw informal tweets and 2.5k annotated informal-formal parallel tweets are publicly accessible\footnote{https://github.com/haryoa/stif-indonesia}.

\subsection{Informal to Formal Data Annotation}

We annotate the informal data into a high form of \textit{Bahasa Indonesia}~\cite{paauw2009malay}. The high form \textit{Bahasa Indonesia} is the standard literary Indonesian omitting the regional varieties and the colloquial words, but including some loan words, online words, and some forms of common conversational Indonesian words. A common theme of these words is English words occurring within a computer or internet-related context (like ``mouse", ``keyboard", ``voucher", "tweet", etc.), which became common with the rise of the Internet before they are standardized into formal  Indonesian. For simplicity, We refer to the high form \textit{Bahasa Indonesia} data as formal data and the opposite as informal data.

We annotate the data by rewriting the given informal text to its formal form. Here are several things that we consider when we annotate the data:
\begin{itemize}

\item \textbf{Punctuation} (e.g.: `Saya bisa admin' $\longrightarrow$ `Saya bisa, admin') (`I can, admin')
\item \textbf{Capitalization} (e.g.: `nama saya haryo' $\longrightarrow$ `Nama saya Haryo') (`my name is Haryo')
\item \textbf{Word order} (e.g.: `Admin, bisa seperti itu kenapa?' $\longrightarrow$ `Admin, kenapa bisa seperti itu?') (`Hey Admin, how is that possible?')
\item \textbf{Colloquial/Shorten Word} (e.g: `gw knp tdk bs' $\longrightarrow$ `Saya kenapa tidak bisa?') (`Why Can't I do this?')
\item \textbf{Affixes or Suffixes} (e.g.: `saya mesenin taxi' $\longrightarrow$ `Saya memesan taxi') (`I ordered a taxi')

\end{itemize}

To ensure the quality of our annotation, we maintain a strict annotation guideline and perform cross-review between annotators. For the experiment purposes, we split the data into train, development, and test. The data is divided into 1922, 214, and 364 for the train, dev, and test data respectively.

\subsection{Data Preprocessing}

We use different pre-processing on each informal and formal data. For informal data, we lowercase the data since the text is irregular, which may hinder the learning process of the model. Then we apply text tranformation when finding more than 2 consecutive characters (e.g.: 'makannn' becomes 'makann'). Finally, we mask the number, account, date, percentage token 
%by using preprocessing open source tool \cite{baziotis-pelekis-doulkeridis:2017:SemEval2}.

\subsection{Data Analysis}

%In this sub-section, we analyze the different forms of data between informal and formal. The purpose of this section is to show changes in informal and formal data changes.

\begin{table}[h!]
    \centering
    \begin{tabular}{l|l|l}
        Stat & Informal data & Formal data \\ \hline
        Number of unique tokens & 5381 & 4036 \\
        Number of tokens & 37040 & 39467 \\
        \hline
        
    \end{tabular}
    \caption{Descriptive statistic of informal and formal data}
    \label{tab:desc-inf-for}
\end{table}

In Table \ref{tab:desc-inf-for}, we show descriptive statistics of the informal and formal data. Where the statistics also include token punctuation and mask token.

The number of unique tokens in informal data is more than formal data. This shows that informal tokens are more varied than formal data. This is because a formal word can be represented with multiple informal tokens. For example, an informal token of 'tidak' ('no') could be 'gak', 'ga', 'nggak', or 'engga'.

\begin{table}[h!]
    \centering
    \begin{tabular}{l|l|l}
        Punctuation & Informal data & Formal data \\ \hline
        . & 1386 & 3488 \\
        , & 1289 & 1821 \\
        ? & 1021 & 1346 \\
        \hline
        
    \end{tabular}
    \caption{Punctuation number of informal and formal data}
    \label{tab:desc-punc}
\end{table}

We found an interesting finding that the number of tokens on the informal data is lower than the formal data, which is the opposite of the unique token number statistics. It shows that there is an increase in the number of punctuation tokens from informal to formal. Table \ref{tab:desc-punc} shows the changes of the punctuation `.', `,', And `?'.
This result makes sense, as people usually pay less attention to punctuation when writing informal text. For example, ``akuu dari awal tahunn gila ga sih" (``I was crazy from the beginning of the year") which becomes `aku sejak awal tahun\textbf{,} gila bukan\textbf{?}' in the formal form. Furthermore, there are informal sentences that do not use `.' at the end of the sentence. We observe that there are 2148 instances in our data where there are no `.' at the end of the sentence.

%\begin{table}[h!]
%    \centering
%    \begin{tabular}{@{}l|l|l|l@{}}
%        Token in Informal Data & Freq. & Token in Formal Data & Freq. \\ \hline
%        saya (`I') & 763 & saya (`I') & 1058 \\
%        min (informal of `admin') & 585 & tidak (`no') & 913  \\
%        ya (`yes') & 554 & admin & 635 \\
%        di (`in') & 510 & sudah ('done') & 545 \\
%        ini (`this') & 425 & bisa ('can') & 510 \\
%        bisa (`can') & 421 & ini ('this') & 482 \\
%        ada (`available') & 324 & di ('in') & 410 \\
%        ga (informal of `tidak';`no') & 272 & yang (`that') & 398 \\
%        mau (`want') & 268 & ada (`available') & 354\\
%        dm (direct message)& 253 & mengapa (`why') & 283 \\
%        yg (informal of `yang';`that') & 232 & terima (`thank') & 277 \\
%        kok (`why' informal) & 229 & kasih (`you') & 277 \\
%        tolong (`help') & 226 & tolong (`help') & 265 \\
%        ke (`to') & 215 & dari (`from') & 253 \\
%        sudah (`done') & 213 & mau (`want') & 240 \\
%        \hline
        
%    \end{tabular}
%    \caption{Top 15 tokens found in informal and formal data. Punctuation and masked token are excluded}
%    \label{tab:desc-rank2}
%\end{table}

\begin{table}[!htb]
    \begin{minipage}{.57\linewidth}
      \centering
        \begin{tabular}{ll}
            Token in Informal Data & Freq. \\ \hline
            saya (`I') & 763 \\
            min (informal of `admin') & 585 \\
            ya (`yes') & 554 \\
            di (`in') & 510 \\
            ini (`this') & 425 \\
            bisa (`can') & 421 \\
            ada (`available') & 324 \\
            ga (informal of `tidak';`no') & 272 \\
            mau (`want') & 268 \\
            dm (direct message)& 253 \\
            yg (informal of `yang';`that') & 232 \\
            kok (`why' informal) & 229 \\
            tolong (`help') & 226 \\
            ke (`to') & 215 \\
            sudah (`done') & 213 \\
            \hline
        \end{tabular}
    \end{minipage}%
    \begin{minipage}{.28\linewidth}
      \centering
        \begin{tabular}{ll}
            Token in Formal Data & Freq. \\ \hline
            saya (`I') & 1058 \\
            tidak (`no') & 913  \\
            admin & 635 \\
            sudah ('done') & 545 \\
            bisa ('can') & 510 \\
            ini ('this') & 482 \\
            di ('in') & 410 \\
            yang (`that') & 398 \\
            ada (`available') & 354\\
            mengapa (`why') & 283 \\
            terima (`thank') & 277 \\
            kasih (`you') & 277 \\
            tolong (`help') & 265 \\
            dari (`from') & 253 \\
            mau (`want') & 240 \\
            \hline
        \end{tabular}
    \end{minipage}
    \caption{Top 15 tokens found in informal and formal data. Punctuation and masked token are excluded}
    \label{tab:desc-rank}
\end{table}

Table \ref{tab:desc-rank} shows the top 15 formal and informal data, which both rank differently according to their word frequency. Tokens `saya', `di', `ini', `sudah', `bisa', `ada', `mau', and `tolong' appear frequently in both informal and formal data. While, several pairs of words in different style are found in the Table, i.e. `min' - `admin', `ga' - `tidak', `kok' - `mengapa', and `yang' - `yg'.

\section{Method}

We address our informal to formal style transfer as a sequence to sequence problem. In this section, we describe our approaches.

\subsection{Dictionary-based Translation}

One naive idea is to simply transform the sequence on a word-level basis with an aid of a dictionary. As a baseline, we developed a simple dictionary-based translator system. We make use of the word-level Indonesian formal-informal dictionary\footnote{https://github.com/ialfina/ID-Kamus-Typo} that has been used in multiple works. This system simply translates an informal word into its formal form if it appears in the dictionary.

\subsection{Phrase-Based Statistical Machine Translation}

Phrase-Based machine translation (PBSMT) has shown to perform well when the dataset is scarce~\cite{lample2018phrase}, as in our case. Hence, we explore using PBSMT as one of the baselines. We employ Moses~\cite{koehn2007moses} to develop our PBSMT system. We use MGIZA~\cite{gao2008parallel} on aligning the phrases.

\subsection{Neural Machine Translation}

The Transformer architecture~\cite{vaswani2017attention} has been state-of-the-art for neural machine translation. Therefore, we explore this architecture as one of our baselines. We employ the standard transformer architecture consisting of 6 layers encoder and decoder. Our model is trained with Marian toolkit~\cite{mariannmt}.

\subsection{Pretrained Language Modeling with GPT-2}

In pre-trained language models, initially, a model (usually Transformer-based) is trained to learn general language modelling. Then, the model will be fine-tuned to the downstream task. Pretrained LM has been shown to adapt to the downstream task without requiring a large dataset, which is suitable for our case. We chose GPT-2 in particular due to its flexibility in fine-tuning downstream generative tasks~\cite{radfordlanguage}.

We first train a GPT-2 based Indonesian language model, which is designed for task generation. Our GPT-2 model is trained on the OSCAR corpus~\cite{suarez2019asynchronous}. Then, we fine-tune our GPT-2 for style-transfer by modifying our parallel corpus into a \texttt{Informal\_sentence <STIF> formal\_sentence} format and train with the modified dataset. (For example, \texttt{cabs kuy <STIF> ayo pergi}). To translate informal sentences, we simply feed the GPT-2 with the informal part and \texttt{<STIF>} tag and let the model complete the sentence. 

\subsection{Synthetic Data Generation}

Commonly, back-translation is used as a synthetic dataset in addition to the parallel corpus. In that case, we need to gather formal Indonesian corpus in customer services domain to be translated to informal text. Unfortunately, formal customer services domain text is uncommon. Therefore, we opt to use a forward-translated synthetic dataset instead. forward-translated data has shown to be beneficial towards to model's performance~\cite{bogoychev2019domain}.

We generate our forward translation model by translating a sample of 5000 tweets with our best-performing model according to fully-supervised learning. We then add this synthetic dataset on top of our original parallel corpus. Finally, we re-train our model with this new dataset composition.

In this experiment, we create an artificial corpus by forward-translating informal text. \citet{hoang2018iterative} has shown that the quality of the synthetic dataset matters. They suggest that iteratively re-construct the synthetic back-translation can improve translation quality. We adopt their finding and attempt an iterative forward-translation: We train our model across different iterations. At the $i$-th iteration, our model is trained with the parallel corpus and additional synthetic forward-translated corpus generated by the model from the $i-1$-th iteration.

\section{Experiment and result}

\begin{table*}[ht!]
    \centering
    \begin{tabular}{ll}
        Method & Examples \\ \hline
        \rowcolor{LightCyan} Input & ku coba resto lain juga sama. jd gmn sih sistemnya? \\ \hline
        \rowcolor{LightCyan} Ground-truth & Saya mencoba \textbf{restoran} lain juga sama. Jadi bagaimana ini sistemnya? \\ 
        \rowcolor{LightCyan} & (I tried other restaurants as well. So how is this the system?) \\ \hline
        Dict-based & ku coba resto lain juga sama. jadi bagaimana sih sistemnya? \\ \hline
        PBSMT & kucoba resto lain juga sama. jadi bagaimana sih sistemnya? \\  \hline
        Transformer & Aku coba \textbf{telpon} lain juga sama. Jadi bagaimana? Apa sistemnya? \\
        & (I tried other phones, it is the same. So how? What is the system?) \\  \hline
        GPT-2 & aku coba resto lain juga sama. jadi bagaimana sistemnya? \\ \hline
        \hline

        \rowcolor{LightCyan} Input & kenapa \textbf{pas} mau login \textbf{kaya} gini terus ya? \\ \hline
        \rowcolor{LightCyan} Ground-truth & mengapa ketika mau login seperti begini terus? \\ 
        \rowcolor{LightCyan} & (why it is always like this whenever I want to log in?) \\ \hline
        Dict-based & kenapa \textbf{pas} mau login \textbf{kaya} gini terus ya? \\ \hline
        PBSMT & kenapa saat mau login seperti ini terus? \\  \hline
        Transformer & kenapa saat mau login seperti ini terus? \\  \hline
        GPT-2 & kenapa saat mau login seperti ini terus? \\ \hline
        
    \end{tabular}
    \caption{some examples of formal text generated from our system.}
    \label{tab:contoh}
\end{table*}

\subsection{Model Benchmark}

\begin{table}[h!]
    \centering
    \begin{tabular}{lr}
        Method & BLEU \\ \hline
        No Modification & 35.32 \\
        Dictionary-Based & 42.11 \\
        PBSMT & \textbf{49.39} \\
        Transformer & 27.50 \\
        GPT-2 Pretrain & \textbf{49.28} \\ \hline \\
    \end{tabular}
    \caption{Style-transfer benchmark with different approaches.}
    \label{tab:bleu-model}
\end{table}

We first answer the question of which approach is the best for low-resource informal-formal Indonesian style transfer. For this purpose, we trained our models with our parallel corpus and evaluate their performance with sacreBLEU~\cite{post2018call}.

As shown in Table \ref{tab:bleu-model}, we present our result of dictionary-based, PBSMT, Transformer, and GPT-2 based approaches. As an additional baseline, we also include \textit{No Modification}, where the input text is unchanged at all.

 Generally, informal Indonesian consists of using colloquial terms of certain words. Therefore, our dictionary-based model performed well with 42.11 BLEU score. One flaw of the dictionary-based model is that it cannot translate words that already exist in the formal dictionary, but have different informal meaning. For example, in the Table~\ref{tab:contoh}, \textit{kaya} (rich, wealthy) is a formal Indonesian word. However, it is also used informally with a different meaning (\textit{kaya} in colloquial Indonesian means 
`similar', or `look like'). The word \textit{pas} also has the same issue (it means 'precise' in formal Indonesian, but means  `whenever' informally). 

Besides, informal Indonesian is also more flexible in the sentence structure. The drawback of this approach is that the model does not consider any word removal or addition. Furthermore, it does not consider words re-alignment. Therefore, it does not perform well when word removal, addition or swap is required. For example, the dictionary-based model cannot remove the suffix 'ya' in the Table~\ref{tab:contoh}.

The result in Table~\ref{tab:bleu-model} shows that the Transformer performed the worst. Consistent with prior work~\cite{lample2018phrase}, we find that the Transformer is incapable of performing well under the extreme low-resource setting. The Transformer performance is worse compared to not modifying the informal text at all. From manual evaluation, we see that the Transformer model often generates output with a different meaning from the source. For example, in the Table~\ref{tab:contoh}, the Transformer output changes the sentence meaning significantly, from complaining about the restaurant to phones. It also alters the sentence structure significantly. We usually find this problem on longer and more complicated sentences.

Our PBSMT approach achieves the best performance of 49.39 BLEU. This result confirms that the PBSMT approach performed better under the extreme low-resource setting. Interestingly, our GPT-2 approach can fine-tune well even with only 2.5k sentence pairs and achieved comparably near-best performance of 49.28 BLEU.

The BLEU scores of GPT-2 and PBSMT approaches are comparable, but the PBSMT is more efficient in terms of computational resource. Therefore, our follow-up experiment will utilize the PBSMT model.

\subsection{Semi-supervised Style-transfer with Forward-Translation}

\begin{figure}
    \centering
    \includegraphics[width=0.9\columnwidth]{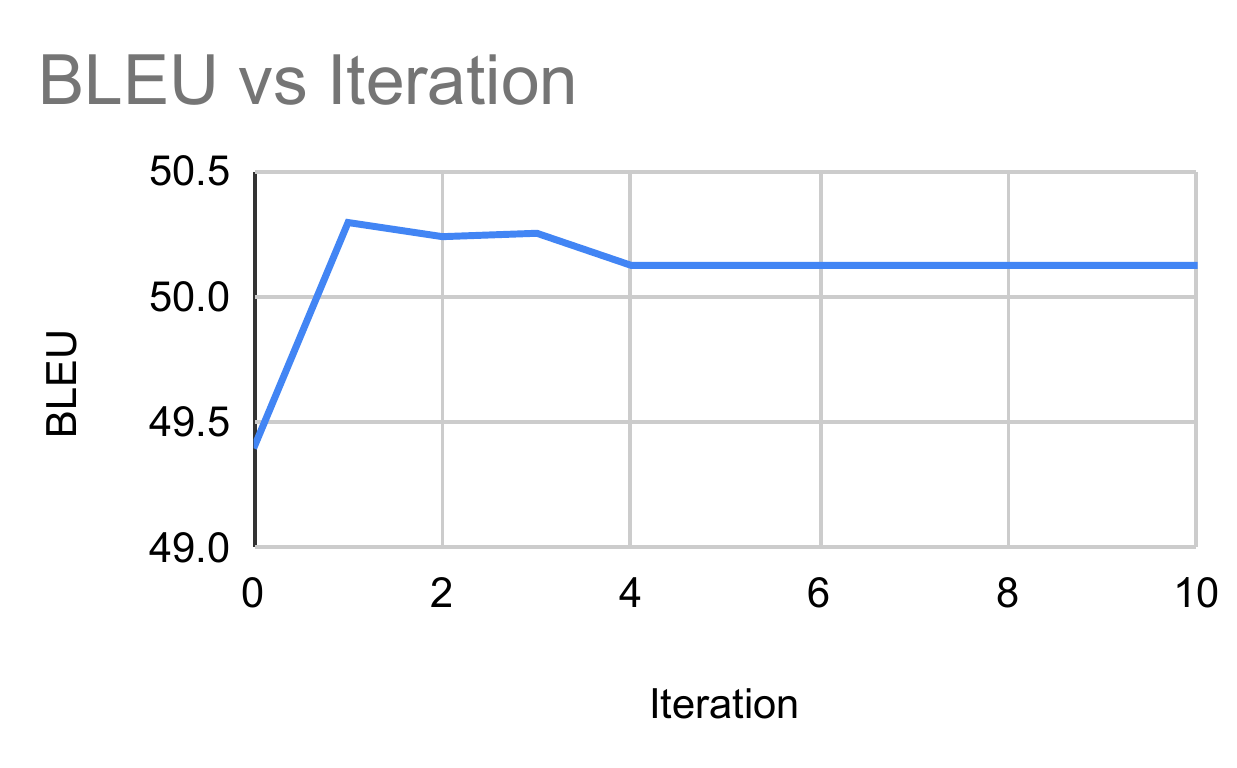}
    \caption{The performance of style-transfer with iterative forward-translation performance in terms of BLEU.}
    \label{fig:iterative}
\end{figure}

We explore adding 2500 forward-translated synthetic sentences on top of the original parallel dataset. We train the model up to 10 iterations. For each iteration, the model is trained normally until convergence. After that, we use the current model to generate the synthetic dataset for the next iteration and then re-start the training again. Model at the iteration 0 is trained only with the original parallel corpus. Our experiment result is shown in the Figure~\ref{fig:iterative}. We use the PBSMT approach as discussed previously.

As shown in in the Figure~\ref{fig:iterative}, adding the synthetic dataset improves the performance to 50.3 BLEU. However, adding more iteration does not seem to improve the performance. This result suggests that the synthetic dataset generated by models at different iterations to be varied enough to produce significantly different performances. The slight decrease in BLEU in later iterations also suggests that the synthetic data introduces noise into the overall training data. We leave the verification of these findings and investigation into their improvements as future work.

\section{Related work}

\citet{rao-tetreault-2018-dear} formulated the problem of formality of text as a style transfer. They released the Grammarly's Yahoo Answers Formality Corpus (GYAFC) dataset, which is parallel data of informal and formal data in the Entertainment \& Music and Family Relationship domain in English. They also tried several sequence-to-sequence approaches to benchmark the data they released. \citet{jhamtani-etal-2017-shakespearizing} also applied sequence-to-sequence approach on style transfer problem. They conducted a style transfer approach by using a modern paraphrase of Shakespeare's play released by \citet{xu-etal-2012-paraphrasing}. We are inspired by their work to apply it into Indonesian informal and formal style. Unfortunately, there are no parallel informal and formal Indonesia data that is ready to be used.

\citet{barik-etal-2019-normalization} aimed to tackle one of the informal Indonesia text problems, namely code-mixed, in which a sentence contains the words in more than one language. They proposed a pipeline solution, i.e. word tokenization, language identification, normalization, and translation. They identified the language of each token in the sentence, then translated the words into Indonesian. 
%Before translating it, they normalized English and Indonesian tokens. 

%Another approach to solve style transfer problems apart from using a sequence to sequence approach is by using a semi-supervised approach. 
\citet{shang-etal-2019-semi} leveraged a semi-supervised approach of style changes by projecting latent space. \citet{yang2018unsupervised} also utilized semi-supervised learning with a generator and discriminator approach to study representations of the latent space of each style.

On the other hand, several works proposed the unsupervised approach. \citet{Luo19DualRL} performed unsupervised style transfer with Dual Reinforcement Learning, where the model was trained adversarially. It used the results of the style classifier as the reward system. \citet{shen2017style} utilized the Variational Auto Encoder (VAE) approach to form latent representation and studied mapping functions to map the latent representation of each style.

\section{Conclusion}

In this paper, we have explored Indonesian informal to formal style-transfer as a low-resource machine translation problem. We build a new parallel informal - formal Indonesian data by annotating tweets from the customer service domain. We conclude that the pretrained GPT-2 and PBSMT approaches achieve the best performance in terms of BLEU. Training with an additional synthetic dataset in the form of forward-translated informal Indonesian text improves the performance.

\section*{Acknowledgement}

This research was supported by the research grant from Universitas Indonesia, namely Publikasi Terindeks Internasional (PUTI) Saintekkes year 2020 no NKB-
2142/UN2.RST/HKP.05.00/2020

\bibliographystyle{plainnat}
\bibliography{coling2020}

\end{document}